\documentclass[10pt,twocolumn,letterpaper]{article}

\usepackage[pagenumbers,datasets]{wacv}
\usepackage[table,dvipsnames]{xcolor}

\definecolor{wacvblue}{rgb}{0.21,0.49,0.74}
\usepackage[pagebackref,breaklinks,colorlinks,allcolors=wacvblue]{hyperref}
\usepackage{amsthm}
\newtheorem{definition}{Definition}
\def\wacvPaperID{272}
\def\confName{WACV}
\def\confYear{2027}

\def\dataset{FOR-species20K}

\begin{document}
\title{Three LiDARs, Twenty-Five Datasets:\\
Cross-Source Generalization in Tree-Species Recognition}

\author{Jae Joong Lee, Bedrich Benes\\
Purdue University, Department of Computer Science\\
305 N University St, West Lafayette, IN 47907, USA\\
{\tt\small lee2161@purdue.edu, bbenes@purdue.edu}
}

\maketitle

\begin{abstract}
\dataset{} is the largest public LiDAR benchmark for tree-species classification, pooling 25 independently collected terrestrial, mobile, and airborne laser-scanning datasets across 33 species. We introduce a multi-resolution, multi-view ConvNeXtV2 model that improves official hidden-test accuracy from 79.5\% to 85.0\%, ranking first at the time of submission. We then examine what this score measures. Because the hidden test is species-balanced, its overall accuracy is more comparable to a macro metric than to accuracy under the public set's long-tailed distribution. Moreover, the official random split places every source dataset in both training and evaluation. A logistic probe recovers source identity from out-of-fold model probabilities with 69\% accuracy, while geometric features identify the source above 90\% within two widely shared species; 18 of the 33 species occur in only one source. Under a source-species-cell holdout, evaluated on 5,566 trees from eight eligible species that retain the same LiDAR-type training support in other sources, the unweighted species mean falls from 82\% to 36\%, and tree-weighted accuracy falls from 95\% to 60\%. Count-matched retraining and experiments with two alternative backbones indicate that this gap is not explained by reduced training support alone, while an external PureForest experiment shows a smaller but consistent grouped-split effect. These results establish the 85\% leaderboard score as strong in-distribution performance, but not as an estimate of source-disjoint generalization, motivating balanced metrics and source-aware evaluation. The code is available at: \url{https://github.com/JaeLee18/FOR20K-lidar-tree-species}
\end{abstract}

\section{Introduction}
\label{sec:intro}
Tree-species recognition from LiDAR asks a direct question: given a laser scan of an individual tree, which species is it? Reliable answers at scale support carbon accounting, biodiversity monitoring, and forest inventory, where manual surveys are expensive to repeat over large areas. \dataset{}~\cite{puliti2025} is the standard benchmark for this task, about 20,000 trees from 25 independently collected source datasets across terrestrial, mobile, and airborne LiDAR.

Point-cloud networks such as PointNet++~\cite{qi2017pointnet++} and DGCNN~\cite{wang2019dgcnn} operate directly on raw 3D points, whereas multi-view methods render each tree as a collection of images processed by image backbones. Image-based methods currently lead on \dataset{}, with the strongest published result reaching 79.5\% overall accuracy using DetailView~\cite{frey2023detailview}. We advance this approach with a multi-resolution, multi-view ConvNeXtV2 classifier: seven geometric views encode depth, point density, and height; a logit-adjusted objective addresses the 33-species long tail; and inference averages predictions from 256- and 512-pixel renders. The resulting model reaches 85.0\% hidden-test accuracy and ranked first on the official leaderboard at the time of submission. We use this high-performing model to investigate how much the official evaluation reflects cross-source species recognition rather than familiarity with the benchmark's source datasets.

A new state-of-the-art invites a new question: what does the leaderboard score measure? Deployment targets forests and sensors that existing collections do not cover. Yet, the official evaluation pairs a species-balanced hidden test with a species-stratified random split that places every source dataset in both the training and test sets. These choices decide what a high score means~\cite{ploton2020spatial, roberts2017cv}. The balanced test moves overall accuracy toward a macro metric: accuracy under equal species weights rather than the public set's long-tailed frequencies. The random split leaves a shortcut open: when species are tied to particular sources, recognizing the source is often sufficient to name the species.

We use this high-performing model to examine what the official evaluation rewards. A logistic probe identifies the source dataset from pooled out-of-fold model probabilities with 69\% accuracy, while handcrafted geometric features identify the source above 90\% within two widely shared species. Under a source-species-cell holdout evaluated on 5,566 trees from eight eligible species that retain the same-LiDAR-type training support in other sources, the unweighted species mean drops from 82\% to 36\%, and tree-weighted accuracy drops from 95\% to 60\%. Count-matched and leave-one-dataset-out retraining indicate that the gap is driven by source structure rather than reduced training support alone. The same qualitative result persists with two alternative backbones, while an external evaluation on PureForest~\cite{gaydon2025pureforest} shows a consistent grouped-split effect as well.

\noindent{Our contributions are as follows.}
\begin{enumerate}
    \item \textbf{A multi-resolution model for LiDAR species recognition.}
    We present a multi-resolution multi-view ConvNeXtv2 classifier for \dataset{}. It improves hidden-test accuracy from 79.5\% to 85\%, and we use it to analyze what the leaderboard rewards (Sec.~\ref{sec:setup}, Tab.~\ref{tab:main}).
    \item \textbf{Balanced-test Interpretation.} We show that the species-balanced hidden test makes official overall accuracy behave much more like a macro metric than accuracy under the true data distribution. Leaderboard scores should therefore be compared with balanced macro metrics on the public labeled set (Sec.~\ref{sec:metric}).
    \item \textbf{Cross-source evaluation.} We show source recognition and cross-source generalization with a source-species holdout that keeps scored species available in the training data. The source is identified from output probabilities at 69\% and at 60\% for species shared across sources. Cross-source evaluation shows a 46\% unweighted species-mean drop and a 35\% tree-weighted drop, with the same trend with two backbones, matched-support and leave-one-dataset-out retraining ablations, and an external benchmark (Sec.~\ref{sec:split} and~\ref{ssec:lso}).
\end{enumerate}

\section{Related Work}\label{sec:related}
\noindent\textbf{Tree species classification from laser scanning.}
Identifying tree species from laser scanning has moved from handcrafted structural and radiometric features~\cite{fassnacht2016review,akerblom2017automatic,terryn2020tree} to deep networks operating on the raw geometry~\cite{hamraz2019deep,seidel2021predicting,briechle2021silvinet,mizoguchi2019individual}. Recent work has moved toward sensor-agnostic models and cross-site generalization for the segmentation and classification of individual trees~\cite{allen2023tree,wielgosz2024segmentanytree,puliti2023forinstance}.

\dataset{}~\cite{puliti2025} brings this direction into the largest public species-classification benchmark, pooling $25$ datasets across three LiDAR types. Complementary to recognition, a parallel line builds 3D tree representations for graphics and simulation, from transformer-based L-system generators~\cite{lee2024latent} to diffusion-based reconstruction of simulation-ready trees from single images~\cite{lee2024treedfusion}.

\noindent\textbf{Point-cloud and multi-view recognition.}
Two model families are commonly used on these data. Point-based networks consume the cloud directly, from PointNet and its hierarchical and graph successors to kernel, transformer, and pure-MLP designs~\cite{qi2017pointnet,qi2017pointnet++,wang2019dgcnn,thomas2019kpconv,zhao2021pointtransformer,wu2022pointtransformerv2,wu2024pointtransformerv3,ma2022pointmlp,qian2022pointnext}, with voxel, sparse-convolution, augmentation, and self-supervised variants~\cite{liu2019pvcnn,graham2018submanifold,choy2019minkowski,hu2020randlanet,li2020pointaugment,pang2022pointmae,yu2022pointbert}.
Multi-view methods instead render the cloud to 2D images for CNNs~\cite{su2015mvcnn,kanezaki2018rotationnet,goyal2021simpleview,choe2022pointmixer}, of which DetailView~\cite{frey2023detailview} is the published work on \dataset{} with the highest accuracy. The benchmark~\cite{puliti2025} finds that every image-based method outperforms every point-cloud method (overall accuracy $0.76$-$0.80$ versus $0.68$-$0.76$), so we build our model on the image family.

\noindent\textbf{Evaluation, leakage, and dataset bias.}
Aggregated benchmarks are known to reward source-specific cues. A classifier can identify the source dataset of an image from its statistics alone~\cite{torralba2011bias}, networks rely on texture and other shortcuts~\cite{geirhos2019imagenet,geirhos2020shortcut}, and recognition that appears solved on a pooled split can collapse across unseen locations and distribution shifts~\cite{beery2018terra,recht2019imagenet,koh2021wilds,gulrajani2021search}.

In spatially autocorrelated data, random cross-validation overestimates accuracy~\cite{ploton2020spatial} and drives reproducibility failures across science~\cite{kapoor2023leakage,leakage_ecology2023}. Grouped or spatially blocked cross-validation is the established remedy~\cite{roberts2017cv,valavi2019blockcv}. The closest tree classification work is the Auto Arborist urban-forest benchmark~\cite{beery2022autoarborist}, which centers geographic domain shift across cities and reports class-averaged recall to counter the long tail. Cross-site evaluation of trees elsewhere points the same way: the IDTreeS competition, spanning three NEON sites report 0.46-0.55 accuracy on its trained sites against 0.07-0.32 on an untrained site~\cite{graves2023data} and PureForest, an aerial-lidar species benchmark drawn from a single nation assigns each forest exclusively to one split~\cite{gaydon2025pureforest}, and WHU-STree spans street trees in two cities~\cite{ding2026whu}. We bring both lessons to the field's flagship multi-sensor LiDAR benchmark. A logistic source classifier measures source information in model outputs, a source-species holdout scores only species with enough training examples outside the held-out source, and macro-F1 accounts for the long tail.

\section{The \dataset{} Benchmark}
\label{sec:bench}

\begin{figure*}[t]
  \centering
  \includegraphics[width=\linewidth]{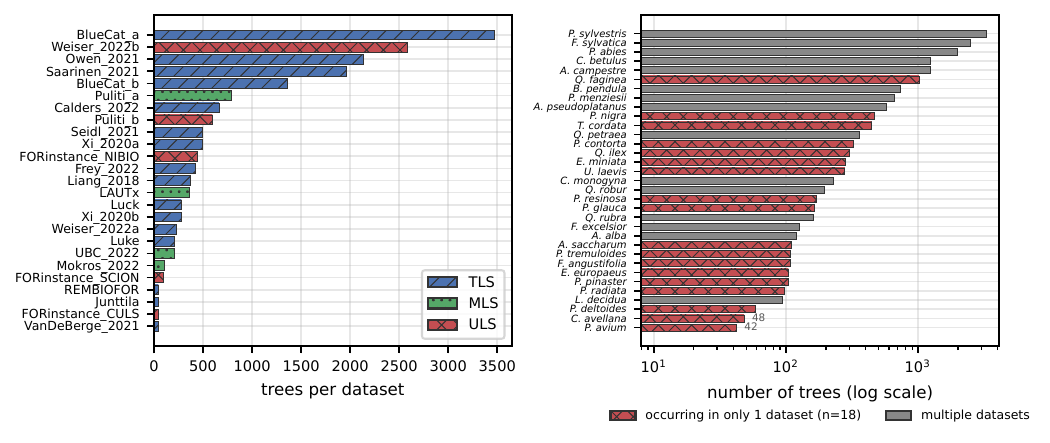}
  \vspace{-0.5cm}
  \caption{Composition of \dataset{}.
  (a) Trees per source dataset, colored by LiDAR type.
  (b) Species frequency on a log scale, with red markers for species that appear in only one dataset.}
  \label{fig:composition}
  \vspace{-0.3cm}
\end{figure*}

\begin{figure*}[t]
  \centering
  \includegraphics[width=\linewidth]{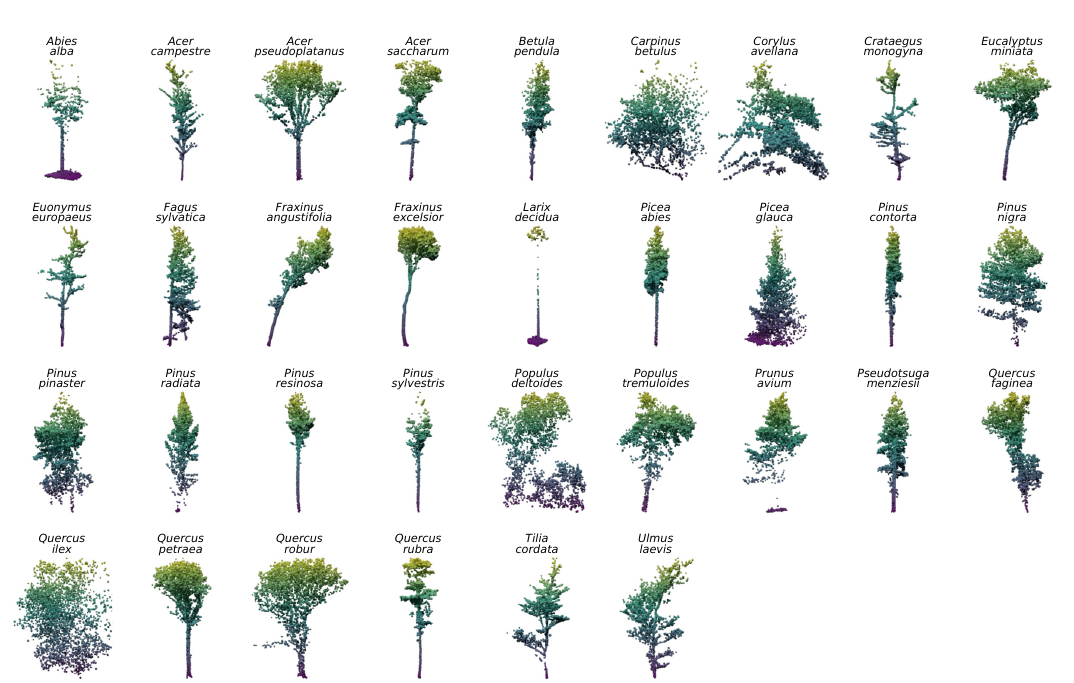}
  \vspace{-0.3cm}
  \caption{The 33 species of \dataset{}.
  A labeled-set tree is shown per species as a raw point cloud with height-coded color.}
  \vspace{-0.3cm}
  \label{fig:data}
\end{figure*}

\dataset{}~\cite{puliti2025} combines 25 independently collected source datasets (Fig.~\ref{fig:composition}) and 33 species (Fig.~\ref{fig:data}).
Source identity is recorded in the metadata's \texttt{dataset} field. Each source was collected using a single LiDAR type, so the source dataset and sensor type are linked. The public release provides a public labeled set of 17,707 trees and a hidden test set of 2,250 trees whose labels are withheld and scored through the Codabench leaderboard~\cite{forspecies20k_leaderboard}. The hidden test set is \emph{species-balanced}, as it samples a similar number of trees per species rather than following the true species distribution.

\label{sec:metric}
Because the hidden test is species-balanced, the official overall accuracy is weighted more toward balanced or macro public-set metrics than toward accuracy under the true data distribution. In the public labeled set, common species account for most of the overall accuracy. In the hidden test, rare and common species receive more similar weights. This is why the public-set overall accuracy of our ensemble, 0.899, is less comparable to the hidden-test score of 0.850 than to its public-set macro-F1 of 0.860. We report macro metrics alongside overall accuracy when using the public labeled set for internal comparison.  The mismatch is not specific to our strongest model: across five image and point-cloud models spanning 0.57-0.89 public-set overall accuracy, accuracy under the true data distribution exceeds macro-F1 in every case, with the gap growing from $+$0.04 for ConvNeXtV2 to $+$0.20 for DGCNN, so macro reporting applies across the benchmark. Complete results are reported in Supplementary Sec.~\ref{app:metricgap} and Table~\ref{tab:metricgap}.

Two benchmark properties matter for the rest of the paper. First, the pooled sources are highly uneven (Fig.~\ref{fig:composition}). They differ in size by two orders of magnitude. Each source uses one LiDAR type, and the species distribution is long-tailed. In particular, 18 of 33 species, about 54\%, occur in only one source dataset. Second, the official evaluation pairs a species-stratified random split with a species-balanced hidden test. The random draw places every source dataset in both training and test, and the balanced test reports accuracy under a species distribution that differs from the true data distribution. These properties are reasonable for a public leaderboard, but they mean that leaderboard accuracy mixes species recognition, class-balance effects, and source familiarity. Section~\ref{sec:method} describes the cross-source evaluation used to separate these effects with our first-place model.

\section{Cross-Source LiDAR Evaluation}
\label{sec:method}
\label{sec:split}
The species-stratified random split places every source dataset in both the training and test. Because each dataset is a single LiDAR type and $18$ of $33$ species occur in only one dataset, recognizing the dataset is often sufficient to name the species, so a random split can reward \emph{source recognition} in place of cross-source generalization. We therefore score held-out source-species pairs only when the species still has training examples in other sources, making the evaluation a test of transfer across sources rather than whether a class was seen at all.

\begin{figure}[t]
  \centering
  \includegraphics[width=0.88\linewidth]{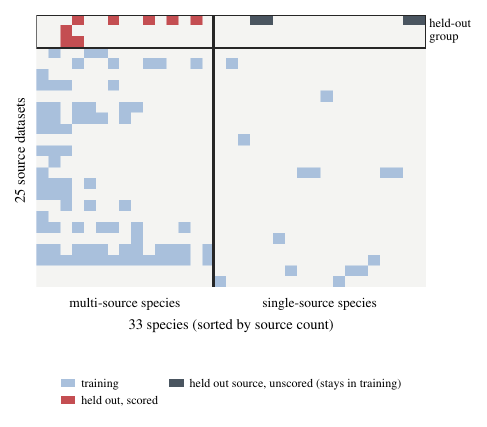}
  \vspace{-0.3cm}
  \caption{Source-species holdout, one fold: 25 sources (held-out group outlined) $\times$ 33 species sorted by source count (divider before the 18 single-source species). Blue = training, red = held out and scored, dark = hidden sources' single-source species.}
  \vspace{-0.5cm}
  \label{fig:protocol}
\end{figure}

\noindent\textbf{Evaluation setup.} We ask whether the model can recognize a species in a source dataset where that source-species was unseen during training. We group source datasets into folds and remove source-species pairs from training for the species that remain available elsewhere (Fig~\ref{fig:protocol}).

\begin{figure*}[hbt]
  \centering
  \includegraphics[width=\linewidth]{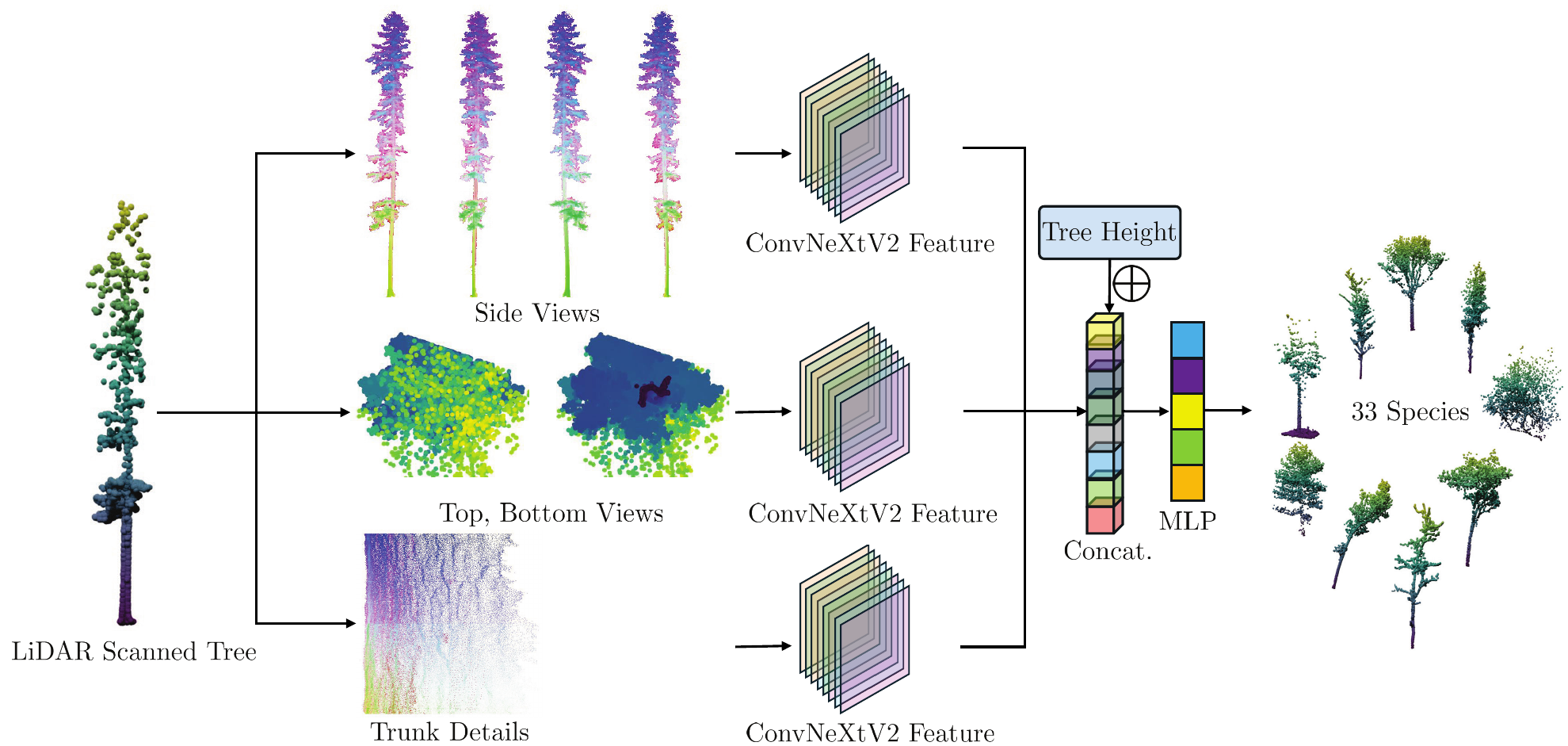}
  \caption{Overview of the multi-view ConvNeXtV2. Seven orthographic renders are split into three view groups. Each group is processed by one ConvNeXtV2-Base backbone~\cite{woo2023convnextv2}, followed by group max-pooling, concatenation with tree height, and a classification head.}
  \label{fig:method}
  \vspace{-0.3cm}
\end{figure*}

\begin{definition}[Cross-source evaluation: a source-species holdout]
\label{def:lso}
We split the 25 source datasets into $K=5$ groups of whole datasets.
Let $G_k$ be the group of source datasets held out in fold $k$.
For that held-out group, the \emph{eligible} species are
\begin{equation}
  S_k = \{\, c : c \text{ appears in some dataset outside } G_k \,\}.
  \label{eq:eligible}
\end{equation}
These are the species that remain present in the training data outside the held-out group. The model is trained on every tree outside the held-out source-species pairs and scored on trees of each $c \in S_k$ that lie in $G_k$. A scored pair, therefore, has no training tree from its own source and species. The same source can still appear in training through other species that are not scored in that fold.
\end{definition}

\noindent The 18 single-source species and any species with no remaining out-of-group training examples stay in training and are not scored as cross-source cases. Four sources contain only single-source species and therefore contribute no eligible pairs, leaving 21 of the 25 sources for scoring. For each eligible tree, we compare its holdout accuracy with its random-split out-of-fold accuracy on the same individual trees, scored under both splits. To separate cross-source failure from cases where the species nearly disappears from training, we report the main gap on species with at least 50 training trees of the held-out LiDAR type outside the held-out group and at least 20 held-out trees. Species whose training data is almost entirely concentrated in $G_k$ are reported separately.

We retrain the model from scratch for each held-out fold and keep every setting identical to Sec.~\ref{sec:setup}. Checkpoint selection follows the same rule in the random split and the cross-source split, so the reported gap is due to the split. We use five folds and report the gap over all of them. Split construction and implementation details are provided in Supplementary Sec.~\ref{app:impl}.

\section{Experimental Setup}
\label{sec:setup}

Our method uses a multi-resolution, multi-view ConvNeXtV2 architecture for tree species classification on \dataset{}. For each tree, we render seven orthographic views, four side views, one top, one bottom, and one trunk close-up at $256 \times 256$ and $512 \times 512$ pixels (Fig.~\ref{fig:method}).
Each view contains three geometric channels representing surface depth, log-transformed point density, and absolute height. We normalize each channel using statistics computed exclusively from the public labeled set. Predictions from the two resolutions are averaged to obtain the final multi-resolution output.

Both resolutions share the three-stream architecture (264.7M parameters in total): three ConvNeXtV2-Base backbones~\cite{woo2023convnextv2} process the view groups, features are max-pooled within each group, concatenated with the tree height, and passed to a two-layer classification head.

To address the long-tailed distribution of the 33 species, we train each model using logit-adjusted cross-entropy~\cite{menon2021longtail}.
For each cross-validation fold, the class prior is estimated from the corresponding training labels.
Models are optimized for 40 epochs using AdamW with a cosine schedule on a single NVIDIA RTX~4090 GPU (about six GPU-hours per fold, and the first three epochs train the head only). Because the channels encode geometry rather than color, augmentation is restricted to horizontal flips and cyclic permutations of the lateral views. The supplementary material provides the details (Tab.~\ref{tab:recipe}).

We perform five-fold cross-validation on the public labeled set and finalize all model and training choices before evaluation on the hidden leaderboard. At inference, predictions are averaged across the five-fold models per resolution and then across the two resolutions.
We will release the source code, trained model weights, and cross-source data splits.

Our multi-resolution model achieves an overall accuracy of 0.850 on the official leaderboard's hidden test set, as shown in Tab.~\ref{tab:main}.
This result improves upon the previously published DetailView result of 0.795~\cite{frey2023detailview}.
Ours also obtains the highest precision, recall, and F1 score on the hidden test set, and records 0.856 for TLS, 0.934 for MLS, and 0.792 for ULS.
At the time of submission, our method ranked first on the official leaderboard.

\begin{table}[t]
  \centering
  \resizebox{\columnwidth}{!}{%
\begin{tabular}{lccccc}
\toprule
Method & Input & OA & Prec. & Rec. & F1 \\
\midrule
DetailView~\cite{frey2023detailview} & Img & 79.5 & 82.3 & 76.7 & 78.0 \\
YOLOv5 & Img & 77.9 & 84.2 & 75.0 & 77.3 \\
SimpleView & Img & 76.2 & 76.9 & 75.5 & 75.6 \\
Ensemble PointNet++~\cite{qi2017pointnet++} & PC & 75.6 & 78.2 & 73.5 & 74.9 \\
MinkNet~\cite{choy2019minkowski} & PC & 73.7 & 79.9 & 70.6 & 72.3 \\
PointMixer & PC & 71.1 & 74.4 & 65.5 & 71.1 \\
PointAugment + DGCNN~\cite{wang2019dgcnn} & PC & 68.3 & 72.5 & 65.7 & 70.3 \\
\midrule
ConvNeXtV2$\times$3 + LA-CE (ours) & Img & 84.0 & 87.0 & 82.0 & 83.0 \\
\textbf{\quad + multi-resolution ensemble (ours)} & Img & \textbf{85.0} & \textbf{89.0} & \textbf{83.0} & \textbf{85.0} \\
\bottomrule
\end{tabular}}

  \vspace{-0.3cm}
  \caption{Official \dataset{} leaderboard~\cite{puliti2025} and our multi-resolution ConvNeXtV2 method.
  All values are percentages, and higher is better.
  Bold marks the best score in each column.}
  \label{tab:main}
  \vspace{-0.3cm}
\end{table}

\begin{table}[t]
  \centering
  \resizebox{\columnwidth}{!}{%
\begin{tabular}{llccccc}
\toprule
Backbone & Rend. & OA & macro-F1 & TLS & MLS & ULS \\
\midrule
DenseNet-201$\times$3 (baseline) & 256 & 0.818 & 0.761 & 0.814 & 0.860 & 0.816 \\
DINOv2-Base$\times$3 & 224 & 0.819 & 0.758 & 0.805 & 0.894 & 0.836 \\
\midrule
ConvNeXtV2-Base$\times$3 (ours) & 256 & 0.886 & 0.843 & 0.879 & 0.934 & 0.894 \\
ConvNeXtV2-Base$\times$3 (ours) & 512 & 0.894 & 0.855 & 0.888 & 0.928 & 0.900 \\
\textbf{\quad + multi-resolution ensemble} & 256$+$512 & \textbf{0.899} & \textbf{0.860} & \textbf{0.893} & \textbf{0.936} & \textbf{0.903} \\
\bottomrule
\end{tabular}}

  \vspace{-0.3cm}
  \caption{Method ablation on the public labeled set with five-fold out-of-fold scores.
  Higher is better.
  The bottom rows show our ConvNeXtV2-Base models and the multi-resolution ensemble.
  Bold marks the best public labeled-set score.
  These scores are not directly comparable to the species-balanced hidden test in Table~\ref{tab:main}.}
  \label{tab:ablation}
  \vspace{-0.3cm}
\end{table}

\noindent\textbf{Ablation Study.}
Table~\ref{tab:ablation} evaluates the contribution of the main design choices using the public labeled set.
The DenseNet-201 baseline trained with class-weighted cross-entropy achieves an out-of-fold overall accuracy of 0.818 and a macro-F1 score of 0.761. With ConvNeXtV2-Base and logit-adjusted cross-entropy, accuracy rises to 0.886 overall and 0.843 macro-F1 pixels. And at 512-pixel input, the model reaches 0.894 and 0.855. The two-resolution ensemble is strongest, at 0.899 overall accuracy and 0.860 macro-F1.

For comparison, we evaluate a self-supervised DINOv2-Base backbone~\cite{oquab2024dinov2}, based on the Vision Transformer architecture~\cite{dosovitskiy2021vit}, using the same training and inference pipeline. The DINOv2 run uses the same seven-view input design, three view-group streams, within-group max pooling, tree-height concatenation, two-layer GELU head, loss, optimizer, schedule, and 40-epoch recipe; only the backbone and its native 224-pixel input size change. It achieves an overall accuracy of 0.819, which is close to that of the DenseNet-201 baseline.
This result indicates that the observed gains cannot be attributed to model capacity alone and instead arise from the combination of the ConvNeXtV2 architecture and the long-tail training objective.

The public labeled-set results in Tab.~\ref{tab:ablation} are out-of-fold estimates under a natural class distribution, whereas Tab.~\ref{tab:main} reports results on the class-balanced hidden test set.
Ensembling with a LightGBM classifier on handcrafted geometric features changed overall accuracy by less than 0.001, so we analyze the multi-resolution ConvNeXtV2 model alone.

\section{Results and Analysis}
\label{sec:results}

\subsection{Random Splits Retain Dataset Cues}
\label{ssec:lso}
\begin{table}[t]
  \centering
  \resizebox{\columnwidth}{!}{%
\begin{tabular}{lcccc}
\toprule
Species & $n$ & random & LSO & $\Delta$ \\
\midrule
\textit{Acer\_pseudoplatanus} & 24 & 0.79 & 0.00 & +0.79 \\
\textit{Betula\_pendula} & 210 & 0.91 & 0.25 & +0.66 \\
\textit{Fagus\_sylvatica} & 44 & 0.93 & 0.34 & +0.59 \\
\textit{Quercus\_robur} & 182 & 0.92 & 0.36 & +0.57 \\
\textit{Picea\_abies} & 1983 & 0.95 & 0.55 & +0.40 \\
\textit{Pinus\_sylvestris} & 3048 & 0.96 & 0.69 & +0.27 \\
\textit{Pseudotsuga\_menziesii} & 43 & 0.93 & 0.70 & +0.23 \\
\textit{Fraxinus\_excelsior} & 32 & 0.12 & 0.00 & +0.12 \\
\midrule
\textbf{Overall} & 5566 & \textbf{0.95} & \textbf{0.60} & \textbf{+0.35} \\
\bottomrule
\end{tabular}}

  \caption{Cross-source accuracy drops on the same trees that are accurate under the random split.
  Each species keeps at least 50 training trees from the same LiDAR type outside the held-out dataset and has at least 20 held-out trees.
  The table reports random-split accuracy and accuracy after the dataset is held out on the same trees.
  A larger $\Delta$ means a larger cross-source drop.
  Bold marks the pooled overall.}
  \label{tab:lso}
  \vspace{-0.5cm}
\end{table}

We first evaluate under the species-stratified random split used for model selection on the public labeled set.
Accuracy is high for many datasets, but some of the easiest cases have very narrow species composition (per-dataset accuracy in the supplementary material).
\emph{Saarinen\_2021}, for example, contains 1958 trees from one species and reaches 99.95\% out-of-fold accuracy, so knowing the dataset already gives the label. Random-split accuracy therefore remains a valid in-distribution score but can mix species recognition with dataset familiarity.

We next ask whether the 33 class probabilities reveal the dataset identity. For each tree, we fit a logistic regression model that predicts the dataset from the species-probability vector. It predicts the correct dataset with 69\% accuracy, compared with a 4\% random-guessing baseline over 25 datasets, and still reaches 60\% when restricted to species observed across multiple datasets. When the species prediction is incorrect, the predicted label is still a species present in the tree's dataset 63\% of the time. The output distribution carries dataset information, and not only for our image model. The same analysis reaches 58\% accuracy on PointNet++ outputs, 66\% on DGCNN outputs, and 67\% on a DINOv2 transformer. The signal is present before any learned model: a classifier on 105 handcrafted geometric features alone recovers the LiDAR type at 96\% and the source dataset at 85\%, with point density, intensity, and returns contributing most. With species held constant, the same features identify which of 12 sources scanned a \emph{Picea\_abies} at 91\% (majority 26\%) and which of 13 a \emph{Pinus\_sylvestris} at 97\%, and removing all 16 acquisition features still leaves the source recoverable at 79\%, so the tested feature-family removals leave the signal largely intact (See the supplementary Sec.~\ref{app:probe} and Tab.~\ref{tab:probe} for the full details). Across eligible species, stronger dataset information in the output is associated with a larger drop in cross-source accuracy.

\begin{figure}[t]
  \centering
  \includegraphics[width=0.92\linewidth]{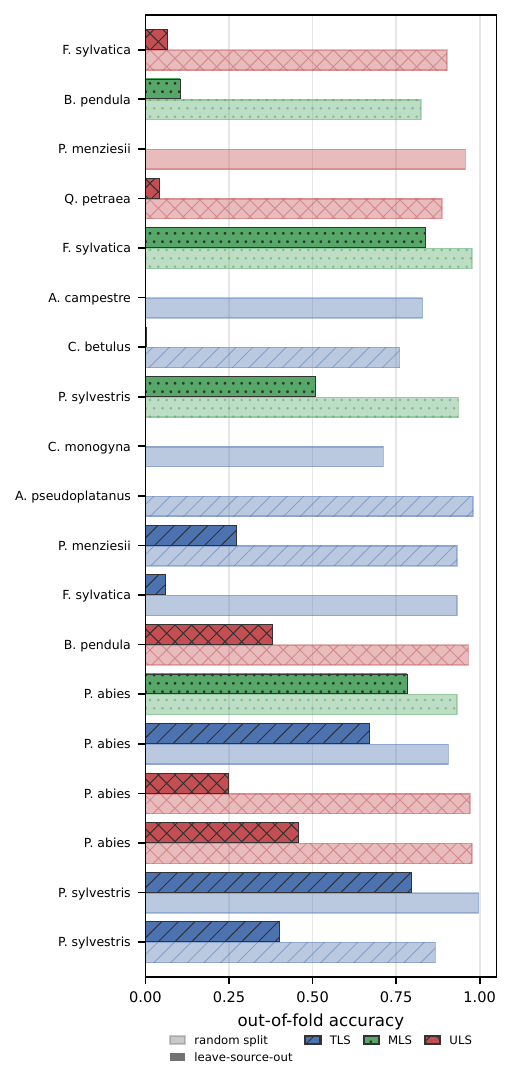}
  \vspace{-0.3cm}
  \caption{Cross-source evaluation exposes the accuracy hidden by the random split.
  Light markers show random-split accuracy on the same trees.
  Solid markers show accuracy after the dataset is held out.
  Rows are colored by LiDAR type and sorted by the number of training trees from the same LiDAR type outside the held-out dataset. Every row shifts left, even when the species remain in training.}
  \label{fig:lso}
  \vspace{-0.3cm}
\end{figure}

The gap appears when one dataset is removed from training while the species remains in other datasets of the same LiDAR type; accuracy follows how many same-type training trees remain (Fig.~\ref{fig:lso}).
\emph{Pinus\_sylvestris} keeps 311 ULS training trees and reaches 68\% accuracy after the largest ULS dataset is removed. \emph{Fagus\_sylvatica} keeps 1647 training trees, but none are ULS after \emph{Weiser\_2022b} is removed, and accuracy drops by 83\%, from 90\% to 7\%.
Same-type examples help but are not sufficient:
\emph{Picea\_abies} retains 408 ULS training trees and still reaches only 25\% accuracy.
Table~\ref{tab:lso} lists the species-level evidence: the largest drops occur in several broadleaf species, and the pooled result remains a 35\% tree-weighted loss, although every scored species keeps same-type training examples outside the held-out dataset. We show detailed fold-level results in the supplementary (Sec.~\ref{app:lso} and Tab.~\ref{tab:lso_full}).

Evaluated on the same held-out trees from the eight eligible species, performance drops sharply when training excludes trees from the same dataset-species group. The unweighted species mean decreases from 82\% under the random split to 36\% under the grouped split, while the tree-weighted score decreases from 95\% to 60\%. This effect is consistent across folds, with drops of 20-61 points, and cannot be attributed to a single unstable species: in a leave-one-species-out analysis, the macro-average drop remains between 41 and 50 points. The two most prevalent eligible species exhibit the same trend, with \emph{Pinus\_sylvestris} and \emph{Picea\_abies} declining by 27 and 40 points, respectively.

Three additional ablations rule out alternative explanations. First, a count-matched experiment retrains one grouped fold after randomly removing, for each held-out (species, LiDAR-type) cell, the same number of training trees as in the grouped-holdout condition. Despite identical per-cell training support, this arm achieves 94\%, comparable to or higher than the 89\% random-split baseline; removing the same number of trees as a coherent source group instead yields 45\%. Thus, the degradation is driven by the source structure of the excluded data rather than by the reduction in sample count. Second, the performance drop is robust to the eligibility threshold and shows no correlation with the remaining per-cell training support. Third, leave-one-dataset-out retraining reduces the pooled drop only modestly, from 42 to 37 points, indicating that the result is not explained solely by the presence of held-out datasets in the training set.

We avoid scoring cases where holding out a dataset removes nearly all training examples of the species. For example, \emph{Acer\_campestre} has 1,228 trees in the held-out dataset and only 2 same-type trees elsewhere. Evaluating that held-out dataset would therefore mostly test recognition of an almost unseen class, not transfer across datasets. Our main estimate avoids this failure mode by scoring only cases with at least 50 same-type training trees outside the held-out dataset, which focuses the measurement on the cross-source shift defined in Sec.~\ref{sec:split}.

The gap also persists when the image backbone is changed. Retrained on all five grouped folds and scored on the support-controlled eligible trees, DenseNet-201 falls from 86\% under the random split to 46\%, and a DINOv2 Vision Transformer falls from 92\% to 49\%; per species, the DINOv2 macro drop is 46\%, matching our first-place model on the same trees. The absolute accuracy depends on the backbone, but the cross-source gap remains.

\subsection{The Split Effect on an Another Benchmark}
\label{ssec:external}

We further evaluate the split effect on PureForest~\cite{gaydon2025pureforest}, an aerial LiDAR species-classification benchmark collected under a single national acquisition program. Of its 18 species, 12 overlap with \dataset{}. We consider 22,993 patches from eight eligible shared species spanning 137 forests and train the same PointNet++ classifier under random and grouped splits. Accuracy decreases from 94\% with a random patch split to 81\% when entire forests are held out. After controlling for the remaining training support, the estimated drop is 9 points, with a forest-level bootstrap confidence interval of 7-18 points.  Grouped accuracy is lower for all eight species (Supplementary Sec.~\ref{app:pureforest}, Tab.~\ref{tab:pf_support} and Tab.~\ref{tab:pureforest}).

This gap is smaller than the 35-point tree-weighted drop observed on \dataset{}. PureForest consists of a single acquisition source, whereas \dataset{} combines 25 independent sources; consistent with this distinction, our feature analysis identifies acquisition properties as an important source of predictive signal. These results suggest that spatial separation contributes to the split effect, which becomes larger when data from heterogeneous sources are combined. PureForest also classifies patches rather than individual trees, and its random split allows adjacent tiles to appear across training and test sets rather than sharing entire acquisition sources. Therefore, we interpret this experiment as directional evidence that the split effect generalizes beyond \dataset{}. Additional details are provided in the supplement (Sec.~\ref{app:pureforest} and Tab.~\ref{tab:pureforest}).

\subsection{Errors in a Dense Mixed Forest}
\label{ssec:failure}
\begin{figure}[t]
  \centering
  \includegraphics[width=0.95\linewidth]{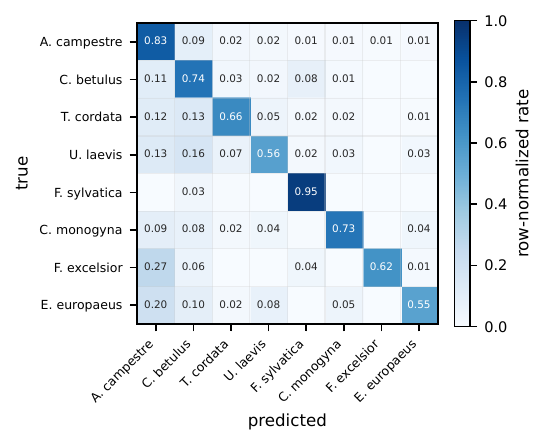}
  \caption{Row-normalized confusion matrix among the co-occurring deciduous broadleaves.
  Rows are true species, and columns are predicted species.}
  \label{fig:confusion}
  \vspace{-0.3cm}
\end{figure}

The remaining public-set errors point to a second difficulty: fine-grained recognition within a dataset already represented in the training folds. This difficulty is most visible in \emph{BlueCat\_a}, a TLS dataset with 3474 trees from a dense mixed-deciduous forest.
Although \emph{BlueCat\_a} is included in the random-split training folds, its out-of-fold accuracy is 73.6\%, and it contributes about half of the public labeled-set errors.

In \emph{BlueCat\_a}, errors occur primarily among co-occurring deciduous species with similar crowns, rather than between conifers and broadleaves. Of these errors, 99\% cross genus boundaries, often favoring frequent classes such as \emph{Acer campestre} and \emph{Carpinus betulus} (Fig.~\ref{fig:confusion}). Across the full public set, cross-genus confusions account for 10.0\% of trees and 88\% of errors, yielding oracle ceilings of 98.6\% overall and 97.9\% balanced accuracy. Collapsing 33 species into 19 genera changes accuracy only from 0.886 to 0.900 and decreases both macro-F1 and balanced accuracy.

These results show that \dataset{} tests fine-grained discrimination among co-occurring species with subtle geometric differences. ConvNeXtV2 improves over DenseNet for every species, with the largest gains on long-tail classes (The full confusion matrix is reported in Sec.~\ref{app:confusion} and Fig.~\ref{fig:confusion_full}), indicating that the improvement is not limited to frequent species.

\section{Conclusion}\label{sec:conc}
We evaluate \dataset{} with a multi-resolution, multi-view model and test how well its score reflects cross-source generalization. Because species balance the hidden test set, official accuracy is close to a macro metric. The species-stratified random split measures in-distribution performance, but can preserve source-specific cues when most trees of a species come from a small number of datasets. Holding out source-species pairs while keeping each species in training reduces the species mean by 46 points and the tree-weighted score by 35 points. PureForest shows the same trend: a forest-grouped split drops 9 points relative to a support-matched random patch split. The smaller drop may reflect its single acquisition program, whereas \dataset{} combines 25 independently collected datasets.

We recommend: (1)~reporting balanced accuracy and macro-F1 alongside overall accuracy; (2)~adding a source-disjoint leaderboard track using the released holdout splits; (3)~reporting per-source results to reveal dependence on single-species sources; (4)~including each species in at least two independent sources, as 18 of 33 species currently appear in only one; and (5)~using the 60\% source-species holdout result, rather than leaderboard accuracy, as the reference for unseen sites when same-type training data are available.

\section*{Funding}
This work is supported by the 2026 Google TPU Builders Research Award and in part by the U.S. National Science Foundation under awards No. 2412928, 2506783 and 2417510. Any opinions, findings, and conclusions or recommendations expressed in this material are those of the author(s) and do not necessarily reflect those of the NSF. This work is based upon efforts supported by the USDA-NIFA grants 2024-67021-42879 ``Improving Forest Management Through Automated Measurement of Tree Geometry'' and 2023-08563 ``Optimizing soybean canopy architecture for efficient light and water use under climate change'' and 2023-68012-38992 grant ``Promoting Economic Resilience and Sustainability of the Eastern U.S. Forests''. The views and conclusions contained herein are those of the authors and should not be interpreted as representing the official policies, either expressed or implied, of the U.S. Government or NRCS. The U.S. Government is authorized to reproduce and distribute reprints for governmental purposes notwithstanding any copyright annotation therein.

{\footnotesize\setlength{\bibsep}{0pt}\linespread{0.90}\selectfont
\bibliographystyle{ieeenat_fullname}
\bibliography{egbib}
}
\clearpage
\newcommand{\secref}[1]{Sec.~\ref{#1}}

\onecolumn

\setcounter{section}{0}
\renewcommand{\theHsection}{A\arabic{section}}
\renewcommand{\thesection}{A\arabic{section}}
\setcounter{table}{0}
\setcounter{figure}{0}
\setcounter{equation}{0}
\renewcommand{\thetable}{A\arabic{table}}
\renewcommand\thefigure{A\arabic{figure}}
\renewcommand{\theHtable}{A.Tab.\arabic{table}}%
\renewcommand{\theHfigure}{A.Abb.\arabic{figure}}%
\renewcommand\theequation{A\arabic{equation}}
\renewcommand{\theHequation}{A.Abb.\arabic{equation}}%

\section*{Supplementary Overview}

The supplementary sections are organized as follows:

\begin{description}
    \item[\secref{app:impl}: Implementation details.] Input representation, network architecture, training objective, optimization, multi-resolution inference, and construction of the source-grouped splits.
    \item[\secref{app:metricgap}: Evaluation metrics.] Overall accuracy, balanced accuracy, and macro-F1 under the natural public-set distribution for all image and point-cloud baselines.
    \item[\secref{app:probe}: Source recognition.] Source information recovered from model-output probabilities, including per-dataset and per-platform results.
    \item[\secref{app:perfold}: Per-fold results.] ConvNeXtV2 out-of-fold performance for each of the five random-split folds.
    \item[\secref{app:perclass}: Per-species results.] Per-species F1 for DenseNet-201 and ConvNeXtV2 across all 33 species.
    \item[\secref{app:confusion}: Error structure.] The full row-normalized confusion matrix for all 33 species.
    \item[\secref{app:lso}: Source-species holdout results.] Complete fold-level results for the source-species-cell evaluation, including the random-split reference and two image backbones.
    \item[\secref{app:pureforest}: PureForest experiment.] Random patch splits versus forest-grouped folds, with support sensitivity and per-species accuracy.
\end{description}

\clearpage

\section{Implementation Details}
\label{app:impl}

\noindent\textbf{Input representation.}
Each tree is rendered to seven $256\times256$ orthographic views: four side views at azimuths $0^\circ$, $90^\circ$, $180^\circ$, $270^\circ$, a top and a bottom view, and one trunk-detail crop.
The three image channels do not encode color; they encode per-pixel surface depth, local point density, and absolute height, so the renders are geometric maps rather than natural photographs.
Channels are normalized with statistics computed over all $17{,}707$ development trees, $\mu=(0.050, 0.028, 0.069)$ and $\sigma=(0.163, 0.109, 0.210)$, not with ImageNet statistics.
Training-time augmentation is restricted to operations that preserve this encoding: a horizontal flip ($p=0.5$) and a cyclic shuffle of the four side views ($p=0.5$, a proxy for a random viewpoint rotation about the vertical axis).
Photometric jitter and vertical flips are excluded, since either corrupts the depth, density, and height semantics.

\noindent\textbf{Network architecture.}
The classifier uses three independent ConvNeXtV2-Base backbones (timm \texttt{convnextv2\_base.fcmae\_ft\_in22k\_in1k}, FCMAE self-supervised pretraining fine-tuned on ImageNet), one per view group.
The four side-view and two top/bottom features are max-pooled within their group; the detail view passes through its backbone directly.
The three $1024$-dimensional group features are concatenated with the scalar tree height and fed to a head of LayerNorm, dropout ($0.3$), a $3073\!\rightarrow\!512$ linear layer, GELU, dropout, and a final $512\!\rightarrow\!33$ linear layer.
The system has $264.7$M parameters.

\noindent\textbf{Training objective.}
Training uses logit-adjusted cross-entropy (Menon et al., 2021): the loss adds $\tau \log \pi_c$ to the logit of class $c$, where $\pi_c$ is the training frequency of class $c$ and $\tau=1$.
This gives rare classes an effective margin without resampling or class weights.
The prior $\pi$ is estimated from each fold's training labels only, and inference uses the raw (unadjusted) logits.

\noindent\textbf{Optimization.}
Table~\ref{tab:recipe} lists the full recipe.
The backbones are frozen for a three-epoch head-only warmup, after which the whole network is fine-tuned under a cosine schedule with AdamW at a $3\times10^{-5}$ backbone and $3\times10^{-4}$ head learning rate.
The effective batch size is $32$ (batch $8$, gradient accumulation $4$) with mixed precision; checkpoint selection uses validation macro-F1.

\noindent\textbf{Multi-resolution ensemble and inference.}
The submitted ensemble averages the softmax outputs of the $256$-pixel model and a second model trained identically at $512$ pixels.
At test time, predictions are averaged over the five fold models and twenty test-time augmentation passes per tree, using the same flip and side-view shuffle as in training; the render resolution at inference matches each model's training resolution.

\noindent\textbf{Support-controlled leave-source-out splits.}
The grouped splits partition the $25$ source datasets into five groups balanced by tree count.
For held-out group $G_k$, the validation set contains every tree in $G_k$ whose species occurs in at least one source outside $G_k$; all remaining trees form the training set.
Each scored species therefore retains training examples from another source.
Species occurring in only one source remain in training and are not scored cross-source.
Training and validation are tree-disjoint within every fold.
The unit of exclusion is a source-species cell: unscored species from the held-out source can remain in training.
Checkpoint selection uses the held-out eligible cells.
The reported LSO accuracies are therefore tuned on the evaluation cells rather than on a separate validation set.

\noindent\textbf{Software and reproducibility.}
Models are implemented in PyTorch with the timm backbone library and trained on a single NVIDIA RTX 4090 (24\,GB), requiring approximately six hours per fold.
Splits are generated with a fixed seed ($42$).
We will release our trained weights, the leave-source-out split files, and the training and evaluation scripts.

\begin{table}[h]
\centering\small
\caption{Training hyperparameters for the ConvNeXtV2 model.}
\label{tab:recipe}
\begin{tabular}{ll}
\toprule
Hyperparameter & Value \\
\midrule
Backbone        & ConvNeXtV2-Base (FCMAE pretrained) \\
Backbones       & 3 (one per view group) \\
Total params    & 264.7M \\
Loss            & Logit-adjusted CE ($\tau=1.0$) \\
Optimizer       & AdamW \\
Backbone LR     & $3\times10^{-5}$ \\
Head LR         & $3\times10^{-4}$ \\
Weight decay    & $10^{-4}$ \\
Scheduler       & Cosine annealing \\
Epochs          & 40 (3 frozen warmup) \\
Effective batch & 32 (batch 8 $\times$ accum 4) \\
Precision       & AMP (mixed fp16/fp32) \\
GPU             & Single NVIDIA RTX 4090 (24 GB) \\
Render size     & $256\times256$ px (7 views/tree) \\
Augmentation    & Horizontal flip, cyclic view-order shuffle \\
CV              & 5-fold stratified (random split) \\
\bottomrule
\end{tabular}
\end{table}

\section{Natural-Distribution and Macro Metrics}
\label{app:metricgap}

Table~\ref{tab:metricgap} compares natural-distribution overall accuracy with balanced accuracy and macro-F1 on the public development set.
Natural-distribution accuracy exceeds macro-F1 for every model, and the gap is largest for the lower-performing point-cloud baselines.
The OA-macro-F1 difference ranges from $0.043$ for ConvNeXtV2 to $0.199$ for DGCNN.

\begin{table}[h]
\centering\small
\caption{Natural-distribution and macro metrics on the public development set, using pooled five-fold out-of-fold predictions.}
\label{tab:metricgap}
\begin{tabular}{lcccc}
\toprule
Model & OA & Bal. acc. & Macro-F1 & OA $-$ Macro-F1 \\
\midrule
ConvNeXtV2 (image) & 0.886 & 0.846 & 0.843 & +0.043 \\
DenseNet-201 (image) & 0.818 & 0.819 & 0.761 & +0.057 \\
DINOv2-Base (image) & 0.819 & 0.759 & 0.758 & +0.061 \\
PointNet++ (point cloud) & 0.573 & 0.433 & 0.460 & +0.113 \\
DGCNN (point cloud) & 0.626 & 0.391 & 0.427 & +0.199 \\
\bottomrule
\end{tabular}
\end{table}

\section{Source Recognition from Model Outputs}
\label{app:probe}

To test whether the output probabilities retain source information (Sec.~6.2 of the main paper), we fit a logistic regression that predicts a tree's source dataset from its $33$-dimensional probability vector.
The inputs are pooled five-fold out-of-fold probabilities rather than raw renders.
Source-classification accuracy is measured with a separate stratified five-fold cross-validation.

Table~\ref{tab:probe} gives the full per-dataset and per-platform breakdown.
Overall source-classification accuracy is $0.691$ ($95\%$ CI $[0.684, 0.698]$), compared with the $0.040$ uniform-chance level.
Randomly permuting the source labels gives $0.194$, equal to the majority-source rate.
Accuracy differs by platform: $0.818$ for TLS, $0.477$ for ULS, and $0.151$ for the four MLS datasets.
The MLS datasets cover nine species and are dominated by \emph{Picea}, \emph{Pinus}, and \emph{Betula}; this restricted composition limits direct comparison with the TLS and ULS results.
For a single-species source, species identity narrows the possible source labels; these sources have high source-classification accuracy.
For the point-cloud baselines, source-classification accuracy is $0.582$ for PointNet++ and $0.655$ for DGCNN.
These baselines were reproduced to obtain out-of-fold probabilities; their \dataset{} species-classification accuracies are the published leaderboard values reported in the main paper (Table~1).

\begin{table}[h]
\centering\small
\caption{Source-recognition accuracy per source dataset, pooled per platform, and overall by backbone, from a logistic source classifier on the model's output probabilities.
Chance is $1/25=0.040$ and higher is better; bold marks the image-modality reference.}
\label{tab:probe}
\setlength{\tabcolsep}{4pt}
\begin{tabular}{llcc}
\toprule
Source dataset & Plat. & $n$ & Source acc. \\
\midrule
\textit{Saarinen\_2021} & TLS & 1958 & 0.999 \\
\textit{Xi\_2020b} & TLS & 278 & 0.993 \\
\textit{FORinstance\_SCION} & ULS & 97 & 0.990 \\
\textit{Luck} & TLS & 280 & 0.982 \\
\textit{Xi\_2020a} & TLS & 488 & 0.980 \\
\textit{BlueCat\_a} & TLS & 3474 & 0.973 \\
\textit{Calders\_2022} & TLS & 661 & 0.935 \\
\textit{BlueCat\_b} & TLS & 1357 & 0.898 \\
\textit{Owen\_2021} & TLS & 2135 & 0.896 \\
\textit{UBC\_2022} & MLS & 204 & 0.809 \\
\textit{Puliti\_b} & ULS & 595 & 0.550 \\
\textit{Weiser\_2022b} & ULS & 2580 & 0.531 \\
\textit{Frey\_2022} & TLS & 426 & 0.225 \\
\textit{Puliti\_a} & MLS & 791 & 0.071 \\
\textit{FORinstance\_CULS} & ULS & 43 & 0.000 \\
\textit{FORinstance\_NIBIO} & ULS & 443 & 0.000 \\
\textit{Junttila} & TLS & 43 & 0.000 \\
\textit{LAUTx} & MLS & 361 & 0.000 \\
\textit{Liang\_2018} & TLS & 366 & 0.000 \\
\textit{Luke} & TLS & 209 & 0.000 \\
\textit{Mokros\_2022} & MLS & 108 & 0.000 \\
\textit{REMBIOFOR} & TLS & 44 & 0.000 \\
\textit{Seidl\_2021} & TLS & 497 & 0.000 \\
\textit{VanDeBerge\_2021} & TLS & 43 & 0.000 \\
\textit{Weiser\_2022a} & TLS & 226 & 0.000 \\
\midrule
\multicolumn{4}{l}{\emph{Per-platform (pooled)}} \\
Pooled (16 datasets) & TLS & 12485 & 0.818 \\
Pooled (4 datasets) & MLS & 1464 & 0.151 \\
Pooled (5 datasets) & ULS & 3758 & 0.477 \\
\midrule
\multicolumn{4}{l}{\emph{Overall, by backbone}} \\
ConvNeXtV2 (image) & All & 17707 & \textbf{0.691} \\
PointNet++ (PC) & All & 17707 & 0.582 \\
DGCNN (PC) & All & 17707 & 0.655 \\
Chance & All & -- & 0.040 \\
\bottomrule
\end{tabular}

\end{table}

\section{Per-Fold Results}
\label{app:perfold}

Table~\ref{tab:perfold} reports ConvNeXtV2 out-of-fold results for all five folds.
The standard deviation across folds is below $0.004$.

\begin{table}[h]
\centering\small
\caption{ConvNeXtV2 out-of-fold accuracy for each fold. Higher is better, and bold marks the five-fold mean.
Internal figures, not comparable to the balanced hidden test.}
\label{tab:perfold}
\setlength{\tabcolsep}{3pt}
\begin{tabular*}{\linewidth}{@{\extracolsep{\fill}}ccccccc@{}}
\toprule
Fold & $n_\text{val}$ & OA & macro-F1 & TLS & MLS & ULS \\
\midrule
0 & 3542 & 0.8905 & 0.8488 & 0.8783 & 0.9486 & 0.9076 \\
1 & 3542 & 0.8862 & 0.8466 & 0.8805 & 0.9301 & 0.8890 \\
2 & 3541 & 0.8870 & 0.8386 & 0.8775 & 0.9233 & 0.9048 \\
3 & 3541 & 0.8820 & 0.8387 & 0.8804 & 0.9156 & 0.8735 \\
4 & 3541 & 0.8865 & 0.8405 & 0.8770 & 0.9532 & 0.8929 \\
\midrule
\textbf{Mean} & -- & 0.8864 & 0.8427 & 0.8787 & 0.9342 & 0.8936 \\
\bottomrule
\end{tabular*}

\end{table}

\section{Per-Species F1 (All 33 Species)}
\label{app:perclass}

\begin{figure}[h]
  \centering
  \includegraphics[height=0.78\textheight]{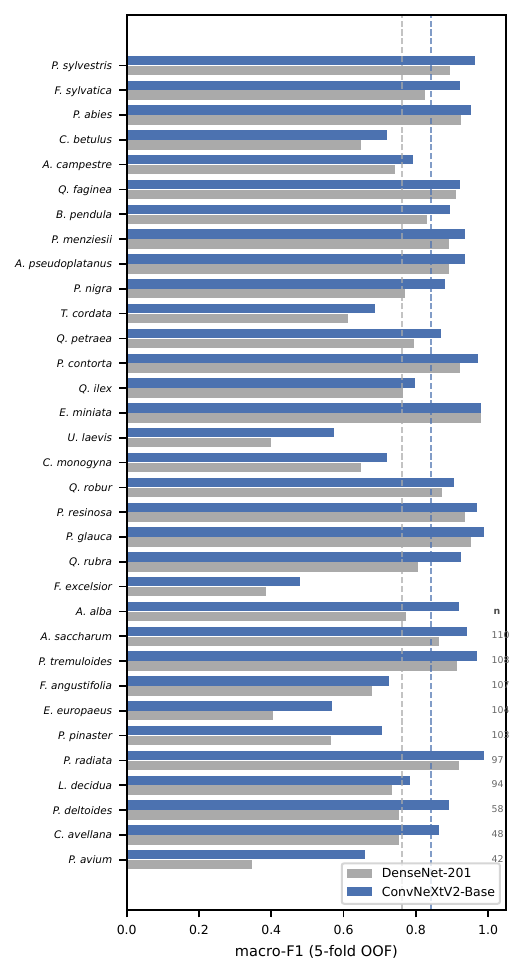}
  \caption{Per-species F1 for DenseNet-201 (gray) and ConvNeXtV2 (blue), sorted by support (rarest at bottom).
  Dashed lines mark each model's mean F1.
  Annotated counts on the right show training support for the rarest ten species.}
  \label{fig:perclass_full}
\end{figure}

Figure~\ref{fig:perclass_full} shows per-species F1 for DenseNet-201 and ConvNeXtV2 across all $33$ species, sorted by training support (rarest at the bottom).
Support counts for the ten rarest species are annotated on the right.
ConvNeXtV2 matches or exceeds the DenseNet-201 F1 for every species; several of the largest differences occur among the rarest classes.
Table~\ref{tab:perclass} gives the exact per-species figures.

\begin{table}[!ht]
\centering\small
\caption{Per-species F1 (higher is better) for DenseNet-201 and ConvNeXtV2-Base, sorted by training support.
$\Delta$ = ConvNeXtV2 $-$ DenseNet.
Bold $\Delta$ indicates a gain $> 0.05$.}
\label{tab:perclass}
\begin{tabular}{lccccc}
\toprule
Species & Support & \multicolumn{2}{c}{F1} & \multicolumn{1}{c}{$\Delta$} \\
\cmidrule(lr){3-4}
 & & DenseNet & ConvNeXtV2 & \\
\midrule
\textit{P. sylvestris} & 3296 & 0.895 & 0.964 & \textbf{+0.069} \\
\textit{F. sylvatica} & 2482 & 0.826 & 0.923 & \textbf{+0.098} \\
\textit{P. abies} & 1983 & 0.924 & 0.951 & +0.027 \\
\textit{C. betulus} & 1243 & 0.648 & 0.720 & \textbf{+0.072} \\
\textit{A. campestre} & 1240 & 0.743 & 0.791 & +0.048 \\
\textit{Q. faginea} & 1014 & 0.911 & 0.921 & +0.010 \\
\textit{B. pendula} & 735 & 0.831 & 0.894 & \textbf{+0.063} \\
\textit{P. menziesii} & 658 & 0.891 & 0.937 & +0.046 \\
\textit{A. pseudoplatanus} & 576 & 0.893 & 0.937 & +0.043 \\
\textit{P. nigra} & 464 & 0.771 & 0.881 & \textbf{+0.110} \\
\textit{T. cordata} & 444 & 0.613 & 0.687 & \textbf{+0.074} \\
\textit{Q. petraea} & 361 & 0.794 & 0.870 & \textbf{+0.077} \\
\textit{P. contorta} & 322 & 0.922 & 0.972 & +0.050 \\
\textit{Q. ilex} & 299 & 0.765 & 0.797 & +0.032 \\
\textit{E. miniata} & 280 & 0.980 & 0.980 & 0.000 \\
\textit{U. laevis} & 278 & 0.398 & 0.574 & \textbf{+0.175} \\
\textit{C. monogyna} & 229 & 0.649 & 0.720 & \textbf{+0.071} \\
\textit{Q. robur} & 195 & 0.873 & 0.906 & +0.033 \\
\textit{P. resinosa} & 168 & 0.935 & 0.970 & +0.035 \\
\textit{P. glauca} & 164 & 0.953 & 0.988 & +0.034 \\
\textit{Q. rubra} & 160 & 0.806 & 0.925 & \textbf{+0.119} \\
\textit{F. excelsior} & 126 & 0.386 & 0.481 & \textbf{+0.095} \\
\textit{A. alba} & 119 & 0.773 & 0.920 & \textbf{+0.148} \\
\textit{A. saccharum} & 110 & 0.865 & 0.942 & \textbf{+0.077} \\
\textit{P. tremuloides} & 108 & 0.913 & 0.968 & \textbf{+0.055} \\
\textit{F. angustifolia} & 107 & 0.678 & 0.726 & +0.048 \\
\textit{E. europaeus} & 104 & 0.404 & 0.567 & \textbf{+0.163} \\
\textit{P. pinaster} & 103 & 0.564 & 0.705 & \textbf{+0.141} \\
\textit{P. radiata} & 97 & 0.919 & 0.990 & \textbf{+0.071} \\
\textit{L. decidua} & 94 & 0.733 & 0.783 & +0.050 \\
\textit{P. deltoides} & 58 & 0.753 & 0.893 & \textbf{+0.139} \\
\textit{C. avellana} & 48 & 0.752 & 0.865 & \textbf{+0.113} \\
\textit{P. avium} & 42 & 0.347 & 0.660 & \textbf{+0.312} \\
\bottomrule
\end{tabular}

\end{table}

\section{Full Confusion Matrix}
\label{app:confusion}

\begin{figure}[h]
  \centering
  \includegraphics[width=0.82\linewidth]{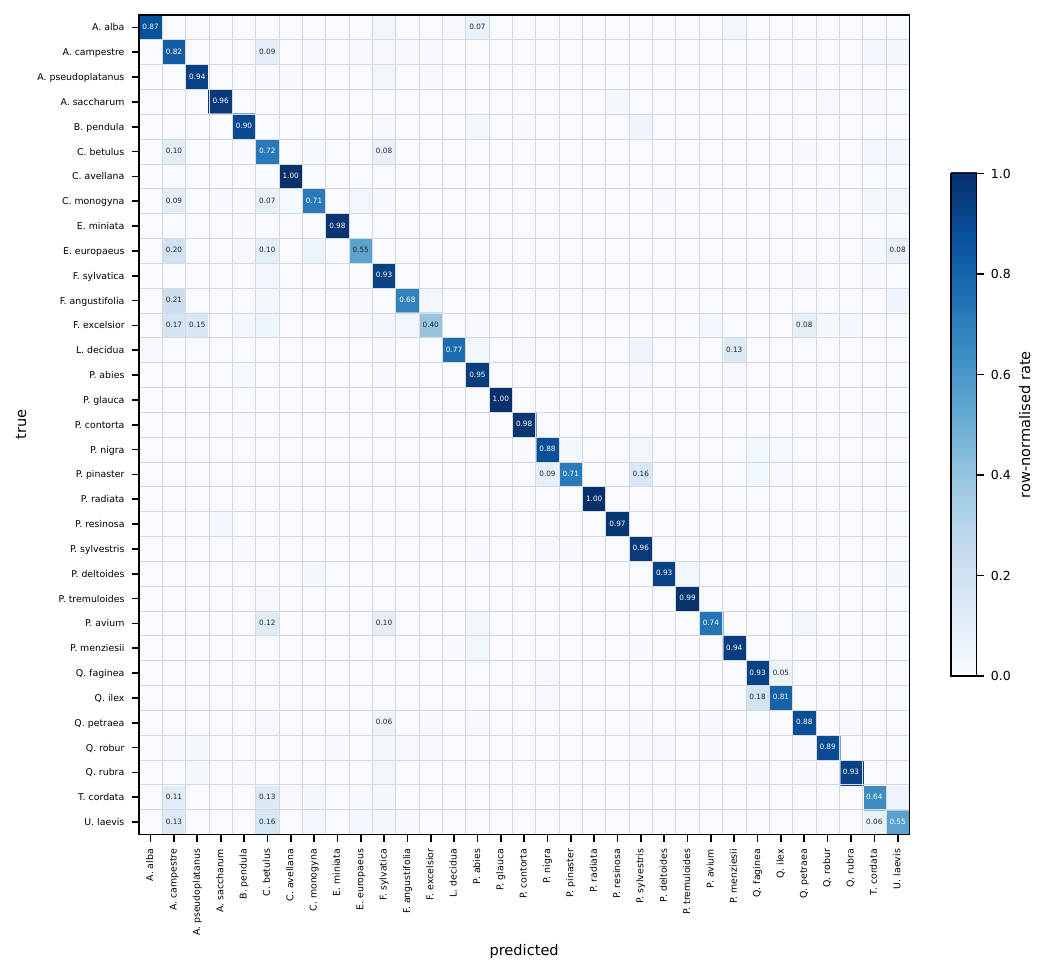}
  \caption{Row-normalized confusion matrix for all 33 species (5-fold OOF, $n{=}17{,}707$).
  Rows are the true species and columns the predicted; each cell is the within-row prediction rate.
  Cells $\geq 0.05$ are annotated.
  Species are in the same order as Table~\ref{tab:perclass}.}
  \label{fig:confusion_full}
\end{figure}

Figure~\ref{fig:confusion_full} shows the ConvNeXtV2 row-normalized confusion matrix over all $33$ species (pooled 5-fold OOF, $n{=}17{,}707$).
Off-diagonal cells $\geq 0.05$ are annotated.
The full matrix has the same structure as the deciduous submatrix in the main paper (Fig.~7).
Conifers and geographically distinctive broadleaves form a near-diagonal block, while most off-diagonal mass lies among the co-occurring deciduous broadleaves in the upper-left block.

\section{Complete Leave-Source-Out Results}
\label{app:lso}

Table~\ref{tab:lso_full} reports support-controlled leave-source-out (LSO) accuracy for every scored species-fold cell.
It includes surviving same-platform training support (\emph{Supp.}), the random-split out-of-fold baseline (\emph{Rand.}), and the LSO accuracies of ConvNeXtV2 and DenseNet-201 retrained on all five grouped folds.
Both backbones lose accuracy under LSO evaluation.
Of the 33 species, 18 occur in a single source and are never cross-source evaluable.
The five grouped folds produce 67 held-out species-platform cells among the multi-source species; 25 cells satisfy both thresholds (at least 20 held-out trees and at least 50 surviving same-platform training trees), covering 8 species and 5,566 held-out trees.
The 42 excluded cells consist of 17 with fewer than 20 held-out trees and 25 with fewer than 50 surviving same-platform training trees.
Seven multi-source species are not scored because their trees are concentrated in a single source.
For these species, holding out that source would remove nearly all training examples and would not isolate cross-source transfer.

\small
\setlength{\tabcolsep}{4pt}
\refstepcounter{table}\label{tab:lso_full}%
\noindent\textbf{Table~\thetable: Complete per-species leave-source-out results.}
\emph{Supp.}\ = same-platform training trees surviving the holdout.
\emph{Rand.}\ = random-split OOF baseline on the same trees.
CNX~LSO = ConvNeXtV2; DNS~LSO = DenseNet (all five grouped folds).
Accuracy columns (\emph{Rand.}, CNX~LSO, DNS~LSO) are higher-is-better.
\vspace{4pt}

\noindent\textbf{Fold 0, held out: }\textit{BlueCat\_a, Junttila, VanDeBerge\_2021}\\[2pt]
\begin{tabular*}{\linewidth}{@{\extracolsep{\fill}}lcccccc@{}}
\toprule
\textbf{Species} & \textbf{Plat.} & \textbf{$n$} & \textbf{Supp.} & \textbf{Rand.} & \textbf{CNX LSO} & \textbf{DNS LSO} \\
\midrule
\textit{Q. robur} & TLS & 40 & 145 & 0.82 & 0.62 & 0.40 \\
\textit{B. pendula} & TLS & 51 & 120 & 0.80 & 0.57 & 0.24 \\
\textit{F. excelsior} & TLS & 32 & 69 & 0.12 & 0.00 & 0.00 \\
\textit{C. monogyna} & TLS & 215 & 14 & 0.71 & 0.00 & 0.00 \\
\textit{C. betulus} & TLS & 1061 & 4 & 0.76 & 0.00 & 0.00 \\
\textit{A. campestre} & TLS & 1228 & 2 & 0.83 & 0.00 & 0.00 \\
\bottomrule
\end{tabular*}
\vspace{6pt}
\noindent\textbf{Fold 1, held out: }\textit{UBC\_2022, Weiser\_2022b}\\[2pt]
\begin{tabular*}{\linewidth}{@{\extracolsep{\fill}}lcccccc@{}}
\toprule
\textbf{Species} & \textbf{Plat.} & \textbf{$n$} & \textbf{Supp.} & \textbf{Rand.} & \textbf{CNX LSO} & \textbf{DNS LSO} \\
\midrule
\textit{P. abies} & ULS & 519 & 408 & 0.97 & 0.25 & 0.14 \\
\textit{B. pendula} & ULS & 7 & 359 & 0.29 & 0.00 & 0.00 \\
\textit{P. sylvestris} & ULS & 170 & 311 & 0.94 & 0.68 & 0.42 \\
\textit{F. excelsior} & ULS & 22 & 3 & 0.09 & 0.00 & 0.00 \\
\textit{A. alba} & ULS & 13 & 0 & 0.69 & 0.00 & 0.00 \\
\textit{A. campestre} & ULS & 10 & 0 & 0.00 & 0.00 & 0.00 \\
\textit{A. pseudoplatanus} & ULS & 46 & 0 & 0.52 & 0.00 & 0.00 \\
\textit{C. betulus} & ULS & 178 & 0 & 0.50 & 0.00 & 0.00 \\
\textit{F. sylvatica} & ULS & 835 & 0 & 0.90 & 0.07 & 0.01 \\
\textit{L. decidua} & ULS & 46 & 0 & 0.72 & 0.04 & 0.02 \\
\textit{P. menziesii} & ULS & 303 & 0 & 0.96 & 0.00 & 0.00 \\
\textit{Q. petraea} & ULS & 337 & 0 & 0.89 & 0.04 & 0.00 \\
\textit{Q. robur} & ULS & 6 & 0 & 0.17 & 0.33 & 0.33 \\
\textit{Q. rubra} & ULS & 86 & 0 & 0.88 & 0.02 & 0.00 \\
\bottomrule
\end{tabular*}
\vspace{6pt}
\noindent\textbf{Fold 2, held out: }\textit{Liang\_2018, Luke, Owen\_2021, Puliti\_b, Weiser\_2022a}\\[2pt]
\begin{tabular*}{\linewidth}{@{\extracolsep{\fill}}lcccccc@{}}
\toprule
\textbf{Species} & \textbf{Plat.} & \textbf{$n$} & \textbf{Supp.} & \textbf{Rand.} & \textbf{CNX LSO} & \textbf{DNS LSO} \\
\midrule
\textit{P. sylvestris} & TLS & 749 & 2048 & 0.87 & 0.40 & 0.27 \\
\textit{F. sylvatica} & TLS & 44 & 1389 & 0.93 & 0.34 & 0.45 \\
\textit{C. betulus} & TLS & 4 & 1061 & 0.25 & 0.00 & 0.00 \\
\textit{A. pseudoplatanus} & TLS & 24 & 506 & 0.79 & 0.00 & 0.00 \\
\textit{P. menziesii} & TLS & 43 & 312 & 0.93 & 0.70 & 0.65 \\
\textit{P. abies} & TLS & 320 & 258 & 0.91 & 0.67 & 0.72 \\
\textit{Q. robur} & TLS & 3 & 182 & 1.00 & 0.67 & 0.67 \\
\textit{Q. rubra} & TLS & 16 & 58 & 0.94 & 0.06 & 0.00 \\
\textit{B. pendula} & ULS & 416 & 46 & 0.97 & 0.38 & 0.26 \\
\textit{A. alba} & TLS & 7 & 45 & 0.71 & 0.57 & 0.29 \\
\textit{L. decidua} & TLS & 1 & 28 & 0.00 & 0.00 & 0.00 \\
\textit{Q. petraea} & TLS & 24 & 0 & 0.83 & 0.33 & 0.00 \\
\bottomrule
\end{tabular*}
\vspace{6pt}
\noindent\textbf{Fold 3, held out: }\textit{Calders\_2022, FORinstance\_NIBIO, LAUTx, Saarinen\_2021}\\[2pt]
\begin{tabular*}{\linewidth}{@{\extracolsep{\fill}}lcccccc@{}}
\toprule
\textbf{Species} & \textbf{Plat.} & \textbf{$n$} & \textbf{Supp.} & \textbf{Rand.} & \textbf{CNX LSO} & \textbf{DNS LSO} \\
\midrule
\textit{A. campestre} & TLS & 2 & 1228 & 0.00 & 0.00 & 0.50 \\
\textit{P. sylvestris} & TLS & 2001 & 609 & 1.00 & 0.79 & 0.64 \\
\textit{P. abies} & ULS & 478 & 564 & 0.98 & 0.46 & 0.35 \\
\textit{B. pendula} & ULS & 39 & 327 & 0.82 & 0.18 & 0.44 \\
\textit{C. monogyna} & TLS & 14 & 215 & 0.71 & 0.14 & 0.07 \\
\textit{Q. robur} & TLS & 33 & 156 & 0.91 & 0.79 & 0.67 \\
\textit{A. alba} & MLS & 4 & 50 & 0.25 & 0.50 & 0.00 \\
\textit{F. excelsior} & TLS & 65 & 39 & 0.63 & 0.02 & 0.00 \\
\textit{A. pseudoplatanus} & TLS & 506 & 24 & 0.98 & 0.00 & 0.00 \\
\textit{F. sylvatica} & MLS & 214 & 0 & 0.98 & 0.84 & 0.76 \\
\textit{L. decidua} & MLS & 19 & 0 & 0.84 & 0.21 & 0.05 \\
\bottomrule
\end{tabular*}
\vspace{6pt}
\noindent\textbf{Fold 4, held out: }\textit{BlueCat\_b, FORinstance\_CULS, Frey\_2022, Mokros\_2022, Puliti\_a, REMBIOFOR, Seidl\_2021}\\[2pt]
\begin{tabular*}{\linewidth}{@{\extracolsep{\fill}}lcccccc@{}}
\toprule
\textbf{Species} & \textbf{Plat.} & \textbf{$n$} & \textbf{Supp.} & \textbf{Rand.} & \textbf{CNX LSO} & \textbf{DNS LSO} \\
\midrule
\textit{P. abies} & MLS & 666 & 115 & 0.93 & 0.78 & 0.42 \\
\textit{F. excelsior} & TLS & 7 & 94 & 0.43 & 0.00 & 0.43 \\
\textit{Q. robur} & TLS & 113 & 72 & 0.95 & 0.12 & 0.42 \\
\textit{F. sylvatica} & TLS & 1389 & 44 & 0.93 & 0.06 & 0.04 \\
\textit{P. menziesii} & TLS & 312 & 43 & 0.93 & 0.27 & 0.12 \\
\textit{Q. rubra} & TLS & 58 & 16 & 1.00 & 0.00 & 0.00 \\
\textit{P. sylvestris} & MLS & 376 & 5 & 0.94 & 0.51 & 0.49 \\
\textit{A. alba} & MLS & 95 & 4 & 0.94 & 0.11 & 0.08 \\
\textit{L. decidua} & TLS & 28 & 1 & 0.82 & 0.04 & 0.04 \\
\textit{B. pendula} & MLS & 222 & 0 & 0.82 & 0.10 & 0.10 \\
\bottomrule
\end{tabular*}

\clearpage

\section{Forest-Level Transfer on PureForest}
\label{app:pureforest}

\subsection{Data and Evaluation Subset}

PureForest (Gaydon and Roche, WACV 2025) contains $135{,}569$ aerial-LiDAR patches from $539$ monospecific forest polygons in $41$ French departments.
Each patch covers $50\,\text{m}\times50\,\text{m}$ and inherits the species label of its forest polygon.
All data were collected under the French national IGN Lidar HD program.

Twelve of the $18$ PureForest species also occur in \dataset{}: \emph{Abies\_alba}, \emph{Fagus\_sylvatica}, \emph{Larix\_decidua}, \emph{Picea\_abies}, \emph{Pinus\_nigra}, \emph{Pinus\_pinaster}, \emph{Pinus\_sylvestris}, \emph{Pseudotsuga\_menziesii}, \emph{Quercus\_ilex}, \emph{Quercus\_petraea}, \emph{Quercus\_robur}, and \emph{Quercus\_rubra}.
The three largest LiDAR archives, \emph{Fagus\_sylvatica} ($15.8$\,GB), \emph{Pinus\_sylvestris} ($17.1$\,GB), and \emph{Quercus\_ilex} ($16.3$\,GB),were excluded to remain within the storage budget.
A species must contain at least five forest polygons for five-fold grouped evaluation.
This removes \emph{Quercus\_rubra}, which has $20$ patches from three forests.
These criteria were fixed before training.
The resulting subset contains $22{,}993$ patches from $137$ forests and eight species.

\subsection{Preprocessing and Training}

Each patch is centered in the horizontal plane, translated so that its lowest point has $z=0$, and subsampled to $4{,}096$ points with seed $42$.
Training examples are further subsampled to $2{,}048$ points, normalized to the unit sphere, randomly rotated about the vertical axis, and perturbed with Gaussian noise ($\sigma=0.01$).
Validation uses deterministic subsampling without augmentation.

Both split conditions use the same PointNet++ single-scale grouping architecture and optimization settings: cross-entropy with label smoothing $0.1$, AdamW with learning rate $10^{-3}$ and weight decay $10^{-4}$, cosine decay over $60$ epochs, batch size $24$, and checkpoint selection by validation macro-F1.
Five models are trained for each split condition.

\subsection{Random and Forest-Grouped Splits}

The random folds are stratified by species at the patch level and ignore forest membership.
Patches from the same forest can therefore occur in training and validation; fold~0 shares $114$ of the $137$ forests.
The grouped folds assign each forest to exactly one fold.
Within each species, a greedy assignment balances patch counts across folds while retaining training forests in every fold.
Training and validation share no forests under this split.
Both conditions produce one out-of-fold prediction per patch, so their predictions are paired on identical examples.

\subsection{Aggregate Results}

Random-split accuracy is $0.936$, compared with $0.815$ for forest-grouped evaluation, giving a patch-weighted drop of $12.1$ points.
The paired patch outcomes yield McNemar counts $b=3083$ and $c=297$ ($p<10^{-300}$).
Because patches from one forest are not independent, the bootstrap samples forests rather than patches.
Across $10{,}000$ forest-level bootstrap samples, the $95\%$ confidence interval for the raw accuracy drop is $[7.0,18.4]$ points.

\subsection{Sensitivity to Remaining Training Support}

The grouped split removes all patches from each validation forest, which can reduce the training support for its species.
Table~\ref{tab:pf_support} reports results after retaining only species-fold cells with at least $S$ grouped-training patches.
The drop is $9.5$ points at $S\geq200$, $9.4$ points at $S\geq400$, and $9.2$ points at $S\geq800$.
Across the $40$ unfiltered species-fold cells, the correlation between the accuracy drop and the retained training fraction is $-0.59$.
The main paper uses the $S\geq800$ result as its support-controlled estimate.

\begin{table}[h]
\centering\small
\caption{Effect of minimum grouped-training support on the PureForest comparison.
At threshold $S$, evaluation includes species-fold cells with at least $S$ training patches in the grouped split.
\emph{Random} and \emph{Grouped} are pooled out-of-fold accuracies on the retained patches; \emph{Drop} $=$ Random $-$ Grouped.}
\label{tab:pf_support}
\caption{Support-controlled PureForest results.
Each row retains species--fold cells with at least $S$ training patches in the grouped split.
\emph{Random} and \emph{Grouped} are pooled five-fold out-of-fold accuracies on the retained patches; \emph{Drop} $=$ Random $-$ Grouped.}
\label{tab:pf_support}
\begin{tabular}{cccccc}
\toprule
$S\geq$ & Cells & Patches & Random & Grouped & Drop \\
\midrule
0   & 40 & 22993 & 0.936 & 0.815 & +0.121 \\
100 & 39 & 22276 & 0.942 & 0.841 & +0.101 \\
200 & 38 & 21974 & 0.941 & 0.846 & +0.095 \\
400 & 37 & 21877 & 0.942 & 0.848 & +0.094 \\
800 & 27 & 21211 & 0.946 & 0.854 & +0.092 \\
\bottomrule
\end{tabular}

\end{table}

\subsection{Per-Species Results}

Table~\ref{tab:pureforest} lists both accuracies for each species.
The grouped score is lower for all eight species.
The largest reductions occur for \emph{Abies\_alba} and \emph{Quercus\_robur}, the two species with the weakest forest-level support.
For \emph{Abies\_alba}, the largest forest contains $717$ of $811$ patches, leaving only $94$ training patches when that forest is held out.
The unweighted species mean assigns equal weight to these low-support cells and gives a $22.8$-point drop.
The main paper reports the support-controlled pooled result instead.

\begin{table}[h]
\centering\small
\caption{PureForest accuracy by species under random patch-level and forest-grouped five-fold evaluation.
Scores are pooled out-of-fold accuracies on identical patches, and $n$ is the patch count.
\emph{Drop} $=$ Random $-$ Grouped, computed before rounding; bold denotes the patch-weighted pooled row.}
\label{tab:pureforest}
\caption{Per-species PureForest results under random patch-level and forest-grouped five-fold splits.
Values are pooled out-of-fold accuracies on the same patches.
$n$ denotes the patch count; \emph{Drop} $=$ Random $-$ Grouped, computed before rounding; bold marks the patch-weighted pooled result.}
\label{tab:pureforest}
\begin{tabular}{lcccc}
\toprule
Species & $n$ & Random & Grouped & Drop \\
\midrule
\textit{A. alba} & 811 & 0.691 & 0.004 & +0.687 \\
\textit{Q. robur} & 447 & 0.902 & 0.441 & +0.461 \\
\textit{P. nigra} & 3198 & 0.873 & 0.694 & +0.179 \\
\textit{Q. petraea} & 3071 & 0.967 & 0.788 & +0.179 \\
\textit{P. menziesii} & 530 & 0.930 & 0.781 & +0.149 \\
\textit{P. abies} & 4074 & 0.931 & 0.840 & +0.091 \\
\textit{P. pinaster} & 7568 & 0.957 & 0.909 & +0.047 \\
\textit{L. decidua} & 3294 & 0.991 & 0.962 & +0.029 \\
\midrule
\textbf{Pooled} & 22993 & \textbf{0.936} & \textbf{0.815} & \textbf{+0.121} \\
\bottomrule
\end{tabular}

\end{table}

\subsection{Relation to \dataset{}}

PureForest measures forest-level transfer within one national acquisition program; it does not include the cross-sensor variation present in \dataset{}.
It also labels fixed-area patches rather than individual trees.
The two comparisons test different forms of group overlap: random PureForest folds share forests and adjacent tiles, whereas random \dataset{} folds share acquisition sources.
The official PureForest train, validation, and test partitions are already forest-disjoint.
The random folds used here are an experimental baseline for quantifying forest overlap, not a replacement for the official split.

\end{document}